\documentclass[journal]{IEEEtran}
\IEEEoverridecommandlockouts
\usepackage[ruled,algoruled]{algorithm2e}
\usepackage{amsmath,amssymb,amsfonts}
\usepackage{balance}
\usepackage{bm}
\usepackage{booktabs}
\usepackage{cite}
\usepackage{enumitem}
\usepackage[T1]{fontenc}
\usepackage{graphicx}
\usepackage{makecell}
\usepackage{mathtools}
\usepackage{multirow}
\usepackage{siunitx}
\usepackage{soul}
\usepackage{subfig}
\usepackage{textcomp}
\usepackage{threeparttable}
\usepackage[color=blue!40]{todonotes}
\usepackage{wrapfig}
\usepackage{xcolor}
\usepackage[]{hyperref}
\soulregister\footnote7
\soulregister\eqref7
\soulregister\cite7
\soulregister\ref7
\DeclareMathOperator*{\argmax}{arg\,max}

\graphicspath{{./}{figures/}{../figures/}}

\begin{document}

\title{%
  Hierarchical Stage-Wise Training of Linked Deep Neural Networks for
  Multi-Building and Multi-Floor Indoor Localization\\
  Based on Wi-Fi RSSI Fingerprinting}%

\author{%
Sihao Li, \IEEEmembership{Graduate Student Member, IEEE}, Kyeong Soo Kim,
\IEEEmembership{Senior Member, IEEE}, Zhe Tang, \IEEEmembership{Graduate
  Student Member, IEEE}, and Jeremy S. Smith, \IEEEmembership{Member, IEEE}%
\\
\thanks{%
  This work was supported in part by the Postgraduate Research Scholarships
  (under Grant PGRS1912001), the Key Program Special Fund (under Grant
  KSF-E-25), and the Research Enhancement Fund (under Grant REF-19-01-03) of
  Xi'an Jiaotong-Liverpool University. This paper was presented in part at
  CANDAR~2023, Matsue, Japan, November 2023.}%
\thanks{%
S. Li and Z. Tang are with the School of Advanced Technology, Xi'an
Jiaotong-Liverpool University, Suzhou, P.R. China (e-mail: [Sihao.Li19,
Zhe.Tang15]@student.xjtlu.edu.cn), and also with the Department of
Electrical Engineering and Electronics, University of Liverpool, Liverpool,
L69 3GJ, U.K. (e-mail: [Sihao.Li, Zhe.Tang]@liverpool.ac.uk).}%
\thanks{%
  K. S. Kim is with the School of Advanced Technology, Xi'an
  Jiaotong-Liverpool University, Suzhou, P.R. China (e-mail:
  Kyeongsoo.Kim@xjtlu.edu.cn).}%
\thanks{%
  J. S. Smith is with the Department of Electrical Engineering and
  Electronics, University of Liverpool, Liverpool, UK (e-mail:
  J.S.Smith@liverpool.ac.uk). }}%
\maketitle
\begin{abstract}
  In this paper, we present a new solution to the problem of large-scale
  multi-building and multi-floor indoor localization based on linked
  neural networks, where each neural network is dedicated to a sub-problem
  and trained under a hierarchical stage-wise training framework. When the
  measured data from sensors have a hierarchical representation as in
  multi-building and multi-floor indoor localization, it is important to
  exploit the hierarchical nature in data processing to provide a scalable
  solution. In this regard, the hierarchical stage-wise training framework
  extends the original stage-wise training framework to the case of
  multiple linked networks by training a lower-hierarchy network based on
  the prior knowledge gained from the training of higher-hierarchy
  networks. The experimental results with the publicly-available
  UJIIndoorLoc multi-building and multi-floor Wi-Fi RSSI fingerprint
  database demonstrate that the linked neural networks trained under the
  proposed hierarchical stage-wise training framework can achieve a
  three-dimensional localization error of 8.19~m, which, to the best of
  the authors' knowledge, is the most accurate result ever obtained for
  neural network-based models trained and evaluated with the full
  datasets of the UJIIndoorLoc database, and that, when applied to a
  model based on hierarchical convolutional neural networks, the proposed
  training framework can also significantly reduce the three-dimensional
  localization error from 11.78~m to 8.71~m.
\end{abstract}


\section{Introduction}
\label{sec:intro}
\IEEEPARstart{W}{ith} the rapid development of sensor technologies and the
increasing popularity of the Internet of Things (IoT), the smart city concept
has been widely adopted in urban areas to improve the quality of life and
enhance the efficiency of urban services. In smart cities, the demand for
localization has increased in various applications, such as asset tracking,
location-based services, and emergency response. The global positioning system
(GPS) is widely used for outdoor localization due to its high accuracy and
global coverage. However, GPS signals are often unavailable indoors due to the
attenuation of signals by walls and ceilings, which makes indoor localization a
challenging problem~\cite{intro_survey_01}.

For an indoor environment, therefore, alternative localization techniques based
on the various sensors equipped on mobile devices (e.g., smartphones) or robots
(e.g., automated guided vehicles (AGVs)), including Wi-Fi, ZigBee, and
Bluetooth Low Energy (BLE) signal sensors and inertial measurement units
(IMUs), have been developed:
With IMUs, for instance, the techniques based on dead reckoning and sensor
fusion have been well-studied and widely used~\cite{IMU_01,IMU_02}.

Among these sensors, Wi-Fi signal sensors are the most popular due to the
ubiquity of Wi-Fi-enabled devices and the wide availability of Wi-Fi access
points (APs)~\cite{STFF}. As for indoor localization techniques based on Wi-Fi,
they can be classified into ranging-based and ranging-free techniques. The
ranging-based techniques, e.g., time of arrival (TOA), time difference of
arrival (TDOA), and angle of arrival (AOA), can achieve high accuracy but
require precise synchronization and calibration, which are costly and
challenging to implement and deploy in practice, especially in large-scale
indoor environments. In contrast, ranging-free techniques are more practical
and cost-effective.

Wi-Fi fingerprinting is a representative example of ranging-free localization.
It starts with the creation of a vector of a pair of a medium access control
(MAC) address and a received signal strength indicator (RSSI) from an AP, which
is measured at a specific location. This vector then becomes a location
fingerprint, which can be utilized to estimate the location of a user/device by
finding the closest match to it in a database of pre-collected location
fingerprints of known locations called reference points (RPs). The Wi-Fi RSSI
fingerprinting technique has been proven to be a practical and effective
solution to the indoor localization problem, as it can estimate locations
within 10-meter-level accuracy even in complex indoor
environments~\cite{intro_survey_02,intro_survey_03}.

Like other techniques based on wireless signal sensors, Wi-Fi RSSI
fingerprinting faces challenges, including signal fluctuations, multi-path
effects, and device and position dependencies in
measurements~\cite{intro_FCLoc,intro_DateRateFP}. To address these challenges,
deep neural networks (DNNs) have been increasingly
utilized~\cite{intro_survey_04,intro_art_01,intro_art_02,intro_art_03,intro_art_04,intro_VirAP,intro_DumbLoc},
which can provide attractive solutions due to their adaptability to a broader
range of conditions with standard architectures and training algorithms. In
particular, DNN-based indoor localization systems provide a unique advantage
over those based on traditional machine learning techniques, as they no longer
need the fingerprint database once trained; the necessary information for
localization is carried by trained DNNs' weights and biases, enabling secure
and energy-efficient indoor localization especially when running on mobile
devices due to there being no exchange of data with the server~\cite{intro_art_04}.

\begin{figure}[!htb]
  \centering%
  \includegraphics[angle=-90,width=.9\linewidth]{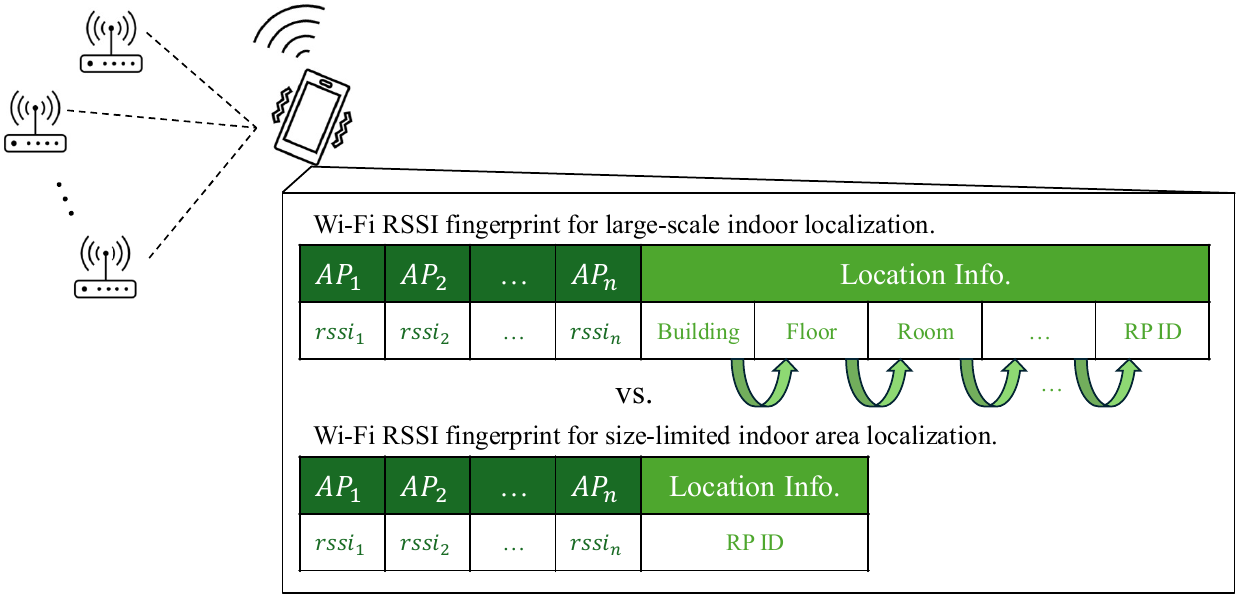}
  \caption{A comparison of the location information of a Wi-Fi RSSI fingerprint
    between multi-building \& multi-floor and single-building \& single-floor
    indoor localization, the former of which shows a hierarchical structure of
    the location.}
  \label{fig:hierarchical}
\end{figure}

When attempting to estimate the location of a user or a device in a large
building complex, such as a shopping mall or a university campus, the
scalability of fingerprinting schemes becomes a significant issue. The
differences in the location information of a Wi-Fi RSSI fingerprint due to the
change of size of the indoor localization volume are illustrated in Fig.~\ref{fig:hierarchical}.
The current state-of-the-art Wi-Fi fingerprinting solutions utilize a
hierarchical approach, i.e., sequentially estimating a location's building,
floor, and floor-level location with a different algorithm tailored for each
task~\cite{intro_art_05}. However, this approach cannot be directly applicable
to DNN-based multi-building and multi-floor indoor localization schemes. As
discussed in~\cite{intro_art_04}, compared to the traditional techniques
proposed in~\cite{intro_art_05}, DNNs for different levels of localization must
be trained separately with multiple sub-databases derived from a common
system-wide database. This process brings considerable challenges of managing
possibly many location fingerprint databases as well as training an equal
number of DNNs.

In this paper, we propose a new training framework, called hierarchical
stage-wise training (HST), based on the extension of the stage-wise
training~\cite{intro_art_06} to multiple linked neural networks (NNs), which
exploits the hierarchical nature of multi-building and multi-floor indoor
localization based on Wi-Fi RSSI fingerprinting. The localization performance
of the linked-DNN model, trained under the proposed training framework,
is evaluated using the UJIIndoorLoc multi-building and multi-floor Wi-Fi
fingerprint database covering three buildings with four or five floors in the
Jaume I University (UJI) campus, Spain~\cite{rela_uji_01}.

The major contributions of our work are three-fold:
\begin{itemize}
  \item First, we present a complete solution to the problem of large-scale
        multi-building and multi-floor indoor localization that can exploit the
        hierarchical nature of the location information of Wi-Fi RSSI fingerprints
        through the HST framework, where the original stage-wise training framework is
        extended to the case of multiple linked NNs.
  \item Second, through the experimental results based on two different DNN
        architectures of a feedforward neural network (FNN) and a convolutional neural
        network (CNN), we demonstrate the flexibility of the proposed solution, which,
        unlike most of the existing approaches, is not limited to a specific DNN
        architecture.
  \item Third, we also provide the state-of-the-art multi-building and multi-floor
        indoor localization performance, which, to the best of the authors' knowledge,
        is the most accurate result obtained for the whole of the UJIIndoorLoc database based on
        a single DNN model.
\end{itemize}

The rest of the paper is organized as follows: Section~\ref{sec:related-work}
reviews the related work. Section~\ref{sec:hst-linked-nns} describes the HST
framework using linked NNs for multi-building and multi-floor indoor
localization based on Wi-Fi RSSI fingerprinting. Section~\ref{sec:exp} presents
the experimental results based on the UJIIndoorLoc database. Finally,
Section~\ref{sec:conc} presents the conclusions of our work.

\section{Related Work}
\label{sec:related-work}

\subsection{Stage-Wise Training}
\label{sec:swt}
Stage-wise training is based on the training of a DNN model in a series of
related sub-tasks called ``stages'' and exploits the evolution, from stage to
stage of (1) the \textit{input domain}, (2) the \textit{output domain}, and (3)
the \textit{training dataset}, of the training process of a
model~\cite{intro_art_06}.

During the training process, the information is injected into the network
gradually so that it focuses on learning the \textit{coarse-scale} properties
of the input data during the early stages, while, during the later stages, it
tries to capture the \textit{finer-scale} and more complete characteristics of
the input data based on the results of the learning at the earlier stages,
which brings a regularization effect and generalizes the learned
representations.

As shown in Fig.~\ref{fig:hierarchical}, when all the sub-tasks for indoor
localization are closely related to one another through a logical chain of
consistent reasoning due to the hierarchical structure of the sensed data,
successful estimation of building-floor-level locations can be highly
beneficial for precise location estimation. Moreover, the stage-wise training
framework can be extended to the case of multiple networks in an evolving
manner to avoid different training difficulties for different tasks and
efficiently handle the problem of large-scale multi-building and multi-floor
indoor localization by taking into account its hierarchical nature.

\subsection{UJIIndoorLoc Database}
\label{sec:uji}
UJIIndoorLoc~\cite{rela_uji_01} is a multi-building and multi-floor Wi-Fi
fingerprint database that served as the official database for the IPIN~2015
competition. The UJIIndoorLoc database covers four or more floors across three
buildings on the UJI campus. It can be used for classification (i.e., the
estimation of the building, floor, and location identifiers (IDs) of an unknown
location) or regression (i.e., the estimation of three-dimensional (3D)
coordinates of an unknown location). The database was constructed in 2013 by
more than 20 different users with 25 Android devices, consisting of 19,937
training records and 1,111 validation/testing records. Its 529 fields include
RSSIs from 520 APs, ranging from -104 to 0~dBm, and location and measurement
information, including building, floor, and space IDs, relative position (i.e.,
inside and outside of the space), and two-dimensional (2D) coordinates (i.e.,
latitude and longitude), user and phone ID, and a timestamp.

\subsection{DNNs for Multi-Building and Multi-Floor Indoor Localization Based on
  Wi-Fi RSSI Fingerprinting}
\label{sec:scalable-DNNs}
For large-scale multi-building and multi-floor indoor localization,
conventional algorithms turned out to be inadequate due to their need for
complex filtering and manual parameter tuning, which is why most of the recent
works in this area rely on deep-learning approaches~\cite{rela_rnn_01}.

A pioneering work in this regard, which is based on a DNN model composed of the
encoder part of a pre-trained stacked autoencoder (SAE) for feature space
dimensionality reduction and a feed-forward classifier for multi-class
classification of building and floor IDs, reports a 92\% success rate in
building and floor classification but does not consider floor-level location
estimation~\cite{intro_art_02}.

To address the scalability issue in large-scale DNN-based multi-building and
multi-floor indoor localization, including floor-level location estimation, a
scalable DNN architecture is proposed in~\cite{intro_art_04}. A DNN model based
on this architecture also consists of the encoder part of a pre-trained SAE and
a feed-forward classifier. Unlike~\cite{intro_art_02}, however, the
feed-forward classifier is trained for \textit{multi-label classification}
using one-hot-encoded output vectors mapped from building, floor, and location
IDs, which can significantly reduce the number of output nodes compared to that
based on multi-class classification. Once trained, three groups of DNN outputs
for building, floor, and location go through customized processing units for
the estimation of building and floor IDs (i.e., the $\argmax$ function) and
floor-level location coordinates (i.e., the \textit{weighted~centroid}),
resulting in a 3D localization error of \SI{9.29}{\meter} with the UJIIndoorLoc
database.

To further reduce the number of output nodes and enable customized training of
building, floor, and location outputs, a single-input and multi-output (SIMO)
DNN architecture based on FNN is proposed in~\cite{rela_dnn_01}, which allows
building/floor classification and floor-level 2D location coordinate regression
through dedicated outputs for each task. The SIMO DNN architecture consists of
common feed-forward hidden layers and three separate feed-forward hidden layers
dedicated to the three groups of outputs for building, floor, and location
estimation. Also, the SAE of~\cite{intro_art_04} is replaced by the stacked
denoising autoencoder (SDAE). For the building and floor outputs, the softmax
activation and categorical cross-entropy loss functions are used for the
multi-class classification of building and floor; for the location output, the
linear activation and mean squared error (MSE) loss functions are used for the
regression of 2D floor-level coordinates. This approach also reduces the number
of output nodes for location estimation to two (i.e., the $x$ and $y$
coordinates) and eliminates the customized processing to convert the
classification results to 2D location coordinates.

In~\cite{rela_cnn_01}, the integration of an SAE and a CNN is explored to
develop a hybrid model named CNNLoc, which prioritizes the success rate for the
estimation of building and floor over the 3D error. Compared to~\cite{intro_art_04}
and~\cite{rela_dnn_01}, CNNLoc requires extensive data pre-processing before
the training phase, including dataset partitioning, creating rectangular areas,
dividing test areas into cell grids, and selecting data based on specific
criteria. CNNLoc could achieve 100\% and 95\% accuracies in building and floor
estimation, respectively, at the expense of a higher 3D error of
\SI{11.78}{\meter} in comparison to~\cite{intro_art_04}.

In~\cite{rela_rnn_01}, a recurrent neural network (RNN) is utilized in
exploiting the hierarchical nature of multi-building and multi-floor indoor
localization. The proposed hierarchical RNN eliminates the need for complex
pre/post-processing of data, requires less parameter tuning, and estimates
locations in a sequential and hierarchical manner, i.e., from building to floor
to location. According to the experimental results, the hierarchical RNN
achieves 100\% and 95.24\% accuracies in building and floor estimation, which
surpasses the existing DNN-based systems mentioned above, and provides a 3D error
of \SI{8.62}{\meter}.

In~\cite{cha22:_wi_fi}, a hierarchical auxiliary deep neural network (HADNN) is
proposed to address the scalability issues in multi-building and multi-floor
indoor localization, where the auxiliary information from the network for a
higher hierarchy is used to train the network for a lower hierarchy together
with the input data. Even though its floor hit rate of 93.15\% and 3D error
performance of \SI{14.93}{\meter}, for the UJIIndoorLoc database, are a bit
higher than those from the state-of-the-art DNN-based systems, it is the first
attempt to apply the hierarchical auxiliary learning~\cite{cha20:_hierar} to
multi-building and multi-floor indoor localization and systematically
investigate its advantages and disadvantages in comparison to other approaches.

\section{Hierarchical Stage-Wise Training of Linked Neural Networks for Wi-Fi
  RSSI Fingerprinting}
\label{sec:hst-linked-nns}
Though the transfer learning was applied to the time-domain adaptation of an
outdated multi-building and multi-floor indoor localization model to a changed
distribution of new fingerprint data in~\cite{TL_indoor_03}, its direct
application to the problem of large-scale multi-building and multi-floor indoor
localization is yet to be fully investigated.

In this section, we present the HST framework that can efficiently train a
model for large-scale multi-building and multi-floor indoor localization, which
are based on not only the knowledge transfer from one domain to another in the
transfer learning but also the evolution of domains through stages in the
stage-wise training. Specifically, we extend the stage-wise training framework
to the case of multiple linked networks to handle the problem of large-scale
multi-building and multi-floor indoor localization by taking into account its
hierarchical nature.

\subsection{Linked-Neural-Network Architecture}
\label{sec:linkednn}
\begin{figure}[!htb]
  \begin{center}
    \includegraphics[angle=-90,width=.9\linewidth]{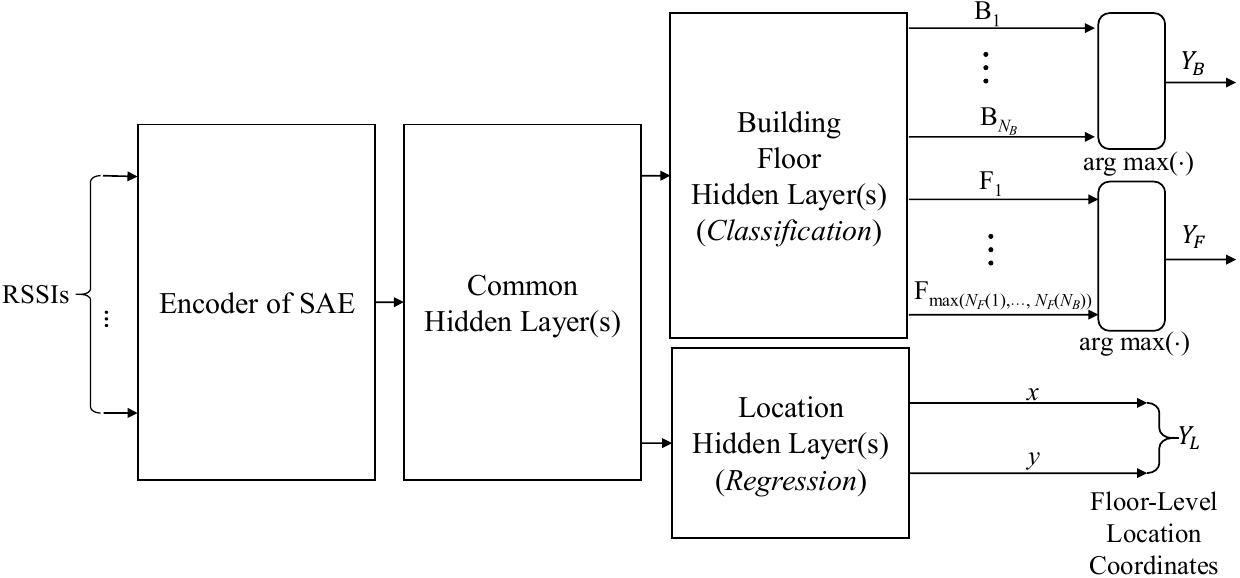}\\
    \vspace{0.25cm}%
    {\footnotesize (a) A simplified version of the hybrid SIMO DNN
      proposed in~\cite{rela_dnn_01}.}\\
    \includegraphics[angle=-90,width=.9\linewidth]{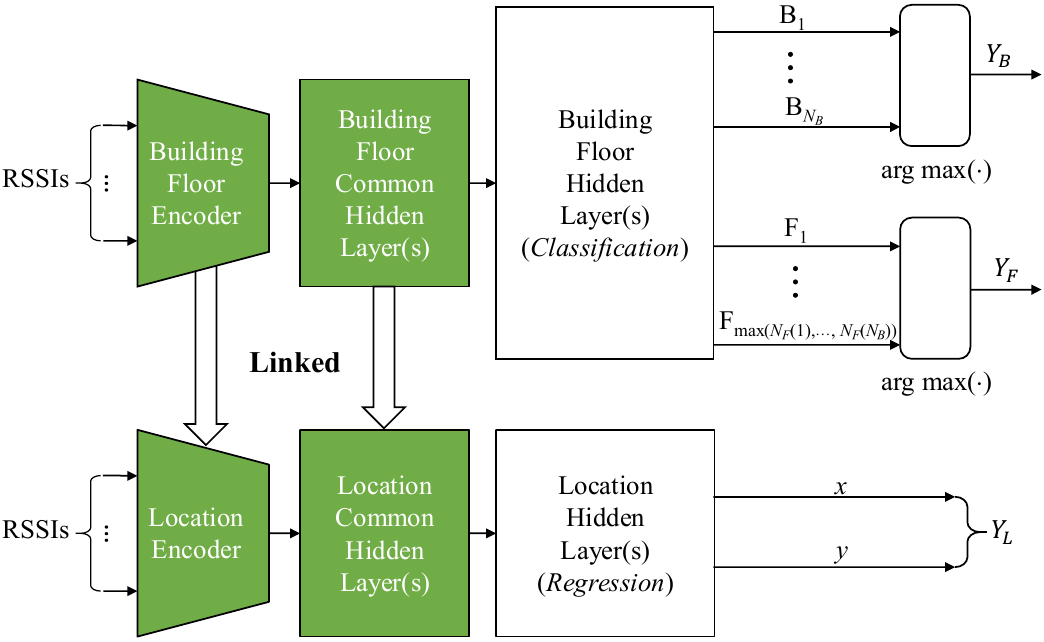}\\
    \vspace{0.25cm}%
    {\footnotesize (b) A modification of (a) based on linked NNs for the HST
      framework.}
  \end{center}
  \caption{DNN architectures for large-scale multi-building and multi-floor
    indoor localization, where the SDAE of~\cite{rela_dnn_01} is replaced by SAE
    for simplicity.}
  \label{fig:dnn_archs}
\end{figure}

Fig.~\ref{fig:dnn_archs}~(a) and (b) show a simplified version of the hybrid
SIMO DNN proposed in~\cite{rela_dnn_01} and its modification based on linked
NNs for the proposed HST framework, respectively, where $N_{B}$ is the number
of buildings in the building complex and
$N_{F}{=}\max\left(N_{F}(1),{\ldots},N_{F}(N_{B})\right)$ with $N_{F}(i)$
($i{=}1,{\ldots},N_{B}$) being the number of floors in the $i$th building.

The hybrid SIMO DNN shown in Fig.~\ref{fig:dnn_archs}~(a) is trained in a
conventional way and serves as a reference in evaluating the localization
performance of the linked NNs trained under the HST framework. Compared to the
\textit{single-input and triple-output} DNN proposed in~\cite{rela_dnn_01}, it
combines the building and the floor outputs into one, which results in a
\textit{single-input and double-output} DNN. As the estimation of the building
and floor IDs is less complex than that of floor-level location, the former is
jointly considered as a multi-label classification problem, while the latter is
formulated as a regression problem. This simplification can also reduce the
training complexity of the linked NNs shown in Fig.~\ref{fig:dnn_archs}~(b)
under the proposed HST framework.

In the linked NNs shown in Fig.~\ref{fig:dnn_archs}~(b), which are specifically
designed for the HST framework, the ``Encoder Part of SAE'' and the ``Common
Hidden Layer(s)'' blocks of the hybrid SIMO DNN are duplicated to the
``Building Floor Encoder'' and ``Building Floor Common Hidden Layer(s)'' blocks
of the DNN for the estimation of building and floor IDs and the ``Location
Encoder'' and ``Location Common Hidden Layer(s)'' blocks of the DNN for the
estimation of floor-level location, respectively. As the \textit{linked blocks}
shown in Fig.~\ref{fig:dnn_archs}~(b) have identical architectures, the learned
parameters of a block can be transferred to the other blocks during the
training under the HST framework.

\begin{figure}[!htb]
  \begin{center}
    \includegraphics[width=.9\linewidth]{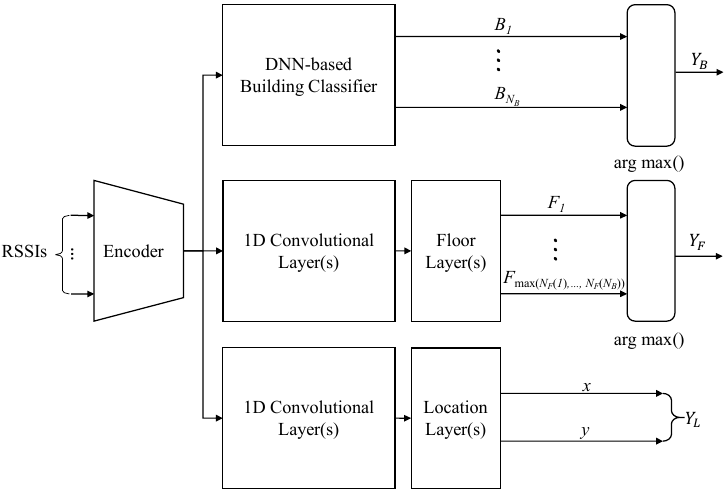}\\
    {\footnotesize (a) Reference model based on CNNLoc~\cite{rela_cnn_01}.}\\
    \vspace{0.2cm}
    \includegraphics[width=.9\linewidth]{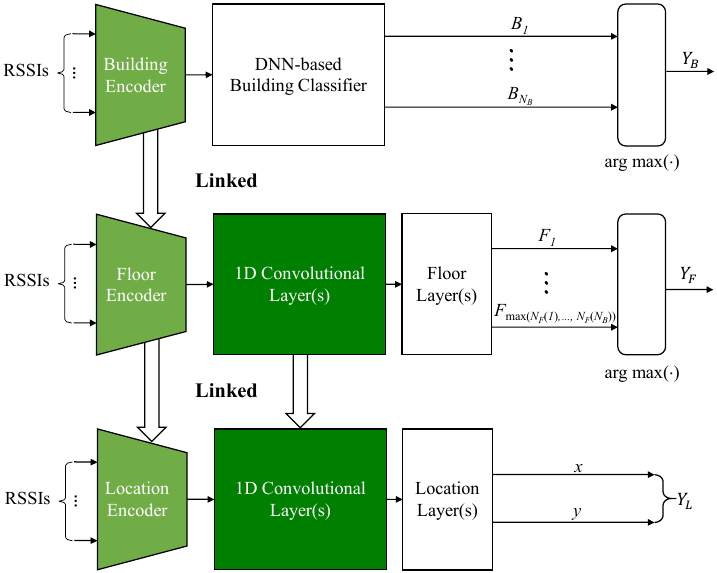}\\
    {\footnotesize (b) A modification of (a) based on linked NNs for the HST
    framework.}
  \end{center}
  \caption{CNN architectures for large-scale multi-building and multi-floor
    indoor localization.}
  \label{fig:cnn_archs}
\end{figure}

Similarly, the CNNLoc proposed in~\cite{rela_cnn_01} can be modified based on
linked NNs for the HST framework as shown in Fig.~\ref{fig:cnn_archs}, where
the ``Encoder'' block is duplicated to the ``Building Encoder'', ``Floor
Encoder'' and ``Location Encoder'' blocks, while the ``1D Convolutional
Layer(s)'' blocks are identical to each other.

\subsection{Hierarchical Stage-Wise Training Framework}
\label{sec:hst}
Given the hierarchical nature of multi-building and multi-floor indoor
localization, the stage-wise training framework~\cite{intro_art_06} is extended
to exploit the evolution of \textit{a parameter set over multiple linked
  networks} as well as the input and output domains and the training set, which
are defined as follows: For stage $s{=}1,2,{\ldots},S$,
\begin{itemize}[noitemsep,topsep=0pt]
  \item Input domain: $\mathcal{X}_{s}$.
  \item Output domain: $\mathcal{Y}_{s}$.
  \item Training dataset:
        \[
          \mathcal{T}_{s} = \left\{\left(\bm{x}_{i},\bm{y}_{i}\right) \,\middle|\,
          i=1, \ldots, n_{s}\right\} \subseteq \mathcal{X}_{s} \times
          \mathcal{Y}_{s}.
        \]
  \item Parameter set:
        \[
          \mathcal{W}_{s} = \left\{{W}_{p,s} \,\middle|\, p \in \mathcal{P}\right\},
        \]
        where ${W}_{p,s}$ are the parameters of a block $p$ of DNNs that can be trained
        as a group at stage $s$.
  \item Learning algorithm: $\mathcal{L}$ is the learning algorithm that takes the
        training dataset and the initial values for the parameter set
        $\mathcal{W}^{init}_{s}=\{{W}_{p,s}^{init}|p{\in}\mathcal{P}\}$ as the input and
        generates the learned parameter set $\mathcal{W}_{s}$ as the output, i.e.,
        \[
          \mathcal{W}_{s} = \mathcal{L}\left(T_{s}, \mathcal{W}^{init}_{s}\right).
        \]
\end{itemize}

Algorithm~\ref{alg:sh} describes the core operations of the HST of linked NNs
for general multi-building and multi-floor indoor localization.
\begin{algorithm}[!htb]
  \caption{HST of linked NNs.}
  \label{alg:sh}
  \footnotesize%
  \KwData{Input domain $\mathcal{X}$; Output domain $\mathcal{Y}$.}
  \KwResult{Parameter set $\mathcal{W}$.}
  Given the number of stages $S$:\\
  ${\qquad}\mathcal{X} = \left\{\mathcal{X}_{1},\mathcal{X}_{2},\,{\ldots}\,,\mathcal{X}_{S}\right\}$\;
  ${\qquad}\mathcal{Y} = \left\{\mathcal{Y}_{1},\mathcal{Y}_{2},\,{\ldots}\,,\mathcal{Y}_{S}\right\}$\;
  ${\qquad}\mathcal{W} = \left\{\mathcal{W}_{1},\mathcal{W}_{2},\,{\ldots}\,,\mathcal{W}_{S}\right\}$\;
  \For{$s=1$ to $S$}{
  Given the number of samples $n_{s}$:\\
  ${\qquad}\mathcal{T}_{s}=\left\{\left(\bm{x}_{i},\bm{y}_{i}\right)\,|\,i=1,\,{\ldots}\,,n_{s}\right\}{\subseteq}\mathcal{X}_{s}{\times}\mathcal{Y}_{s}$\;
  Given the number of blocks $P$:\\
  ${\qquad}\mathcal{W}_{s}=\left\{{W}_{1,s},{W}_{2,s},\,{\ldots}\,,{W}_{P,s}\right\}$\;
  \For{$p=1$ to $P$}{
  \eIf{$s=1$}{${W}_{p,s}^{init} := \text{random or pre-trained}$}{
  \eIf{Block $p$ has a linked block at stage $s{-}1$.}{${W}_{p,s}^{init} := {W}_{p,s-1}$}{${W}_{p,s}^{init} := \text{random or pre-trained}$}
  }
  }
  $\mathcal{W}_{s}=\mathcal{L}\left(T_{s},\mathcal{W}^{init}_{s}\right)$\;
  Save and frozen $\mathcal{W}_{s}$.
  }
\end{algorithm}

\begin{table}[!htb]
  \centering%
  \caption{Blocks of the linked-DNN model shown in Fig.~\ref{fig:dnn_archs}~(b)
    for the HST framework.}
  \label{tab:dnn_exp}
  \begin{tabular}{clc}
    \toprule
    Stage                           & \multicolumn{1}{c}{Block}                                & Symbol     \\
    \midrule
    \multirow{3}*{\textbf{Stage 1}} & Building Floor Encoder                                   & $E_{B\!F}$ \\
                                    & Building Floor Common Hidden Layer(s)                    & $H_{B\!F}$ \\
                                    & Building Floor Hidden Layer(s) (\textit{Classification}) & $C$        \\
    \cmidrule{2-3}
    \multirow{3}*{\textbf{Stage 2}} & Location Encoder                                         & $E_{L}$    \\
                                    & Location Common Hidden Layer(s)                          & $H_{L}$    \\
                                    & Location Hidden Layer(s) (\textit{Regression})           & $R$        \\
    \bottomrule
  \end{tabular}
\end{table}

Applying Algorithm~\ref{alg:sh} to the linked-DNN model shown in
Fig.~\ref{fig:dnn_archs}~(b), based on the blocks described in
Table~\ref{tab:dnn_exp}, we can train them under the proposed HST framework as
follows:

\noindent%
\textbf{Stage 1}:
\begin{itemize}[noitemsep,topsep=0pt]
  \item Input domain: $\mathcal{X}_{1}$ is the domain of scaled RSSI values, i.e.,
        \begin{equation*}
          \begin{aligned}
            \mathcal{X}_{1} = & \big\{\left(\text{RSSI}_{1},\ldots,\text{RSSI}_{N}\right) \,\big|     \\
                              & \quad \text{RSSI}_{i} \in \mathbb{R} \text{ for } i=1,\ldots,N\big\},
          \end{aligned}
        \end{equation*}
        where $N$ is the number of APs in the building complex.
  \item Output domain: $\mathcal{Y}_{1}$ is the domain of concatenated one-hot-encoded
        categorical variables for building and floor IDs, i.e.,
        \begin{equation*}
          \begin{aligned}
             & \mathcal{Y}_{1} = \big\{\left(b_{1},\ldots,b_{N_{B}},f_{1},\ldots,f_{N_{F}}\right) \,\big|          \\
             & \quad b_{i} \in \{0,1\} \text{ for } i=1,\ldots,N_{B} \text{ and } \sum_{i=1}^{N_{B}}b_{i}=1,       \\
             & \quad f_{j} \in \{0,1\} \text{ for } j=1,\ldots,N_{F} \text{ and } \sum_{j=1}^{N_{F}}f_{j}=1\big\}.
          \end{aligned}
        \end{equation*}
  \item Training dataset:
        \[
          \mathcal{T}_{1} = \left\{\left(\bm{x}_{i}, \bm{y}_{i}\right) \,\middle|\,
          i=1, \ldots, n_{s}\right\} \subseteq \mathcal{X}_{1} \times
          \mathcal{Y}_{1}.
        \]
  \item Parameter set:
        \[
          \mathcal{W}_{1} = \left\{{W}_{E_{B\!F},1}, {W}_{H_{B\!F},1},
          {W}_{C,1}\right\}.
        \]
  \item Initialization:
        \begin{align*}
           & {W}_{E_{B\!F},1}^{init} := \text{pre-trained\footnotemark}, \\
           & {W}_{H_{B\!F},1}^{init} := \text{random},                   \\
           & {W}_{C,1}^{init}        := \text{random}.
        \end{align*}
        \footnotetext{The ``Building Floor Encoder'' is from the SAE that is
          pre-trained as described in~\cite[Fig. 1]{SAE_train_01}.}
  \item Learning algorithm:
        \[
          \mathcal{W}_{1} = \mathcal{L}\left(\mathcal{T}_{1}, \mathcal{W}^{init}_{1}\right).
        \]
\end{itemize}

\noindent%
\textbf{Stage 2}:
\begin{itemize}[noitemsep,topsep=0pt]
  \item Input domain: $\mathcal{X}_{2}=\mathcal{X}_{1}$.
  \item Output domain: $\mathcal{Y}_{2}$ is the domain of scaled floor-level 2D
        coordinates, i.e.,
        \begin{equation*}
          \mathcal{Y}_{2} = \left\{(x,y) \,\middle|\, x,y \in \mathbb{R}\right\}.
        \end{equation*}
  \item Training dataset:
        \[
          \mathcal{T}_{2} = \left\{\left(\bm{x}_{i},\bm{y}_{i}\right) \,\middle|\,
          i=1, \ldots, n_{s}\right\} \subseteq \mathcal{X}_{2} \times
          \mathcal{Y}_{2}.
        \]
  \item Parameter set:
        \[
          \mathcal{W}_{2} = \left\{{W}_{E_{L},2}, {W}_{H_{L},2}\,
          {W}_{R,2}\right\}.
        \]
  \item Initialization:
        \begin{align*}
           & {W}_{E_{L},2}^{init} := {W}_{E_{B\!F},1}, \\
           & {W}_{H_{L},2}^{init} := {W}_{H_{B\!F},1}, \\
           & {W}_{R,2}^{init}     := \text{random}.
        \end{align*}
  \item Learning algorithm:
        \[
          \mathcal{W}_{2} = \mathcal{L}\left(\mathcal{T}_{2}, \mathcal{W}^{init}_{2}\right).
        \]
\end{itemize}

\begin{figure}
  \begin{center}
    \includegraphics[angle=-90,width=.9\linewidth]{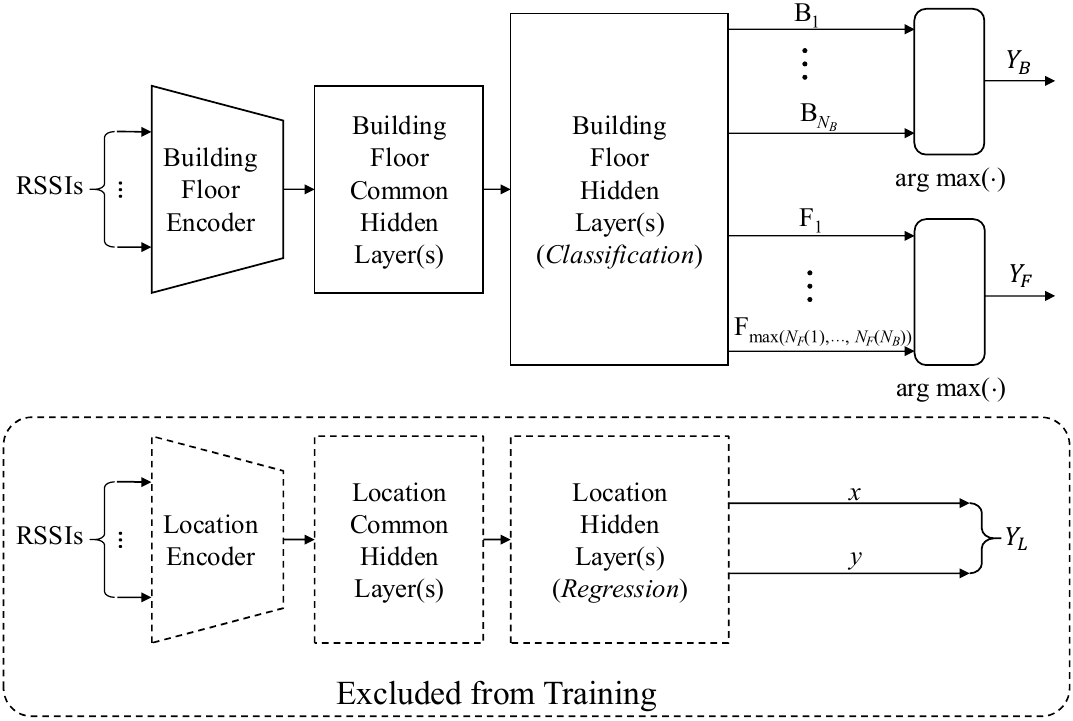}\\
    \vspace{0.3cm}%
    {\footnotesize (a) Stage 1}\\
    \includegraphics[angle=-90,width=.9\linewidth]{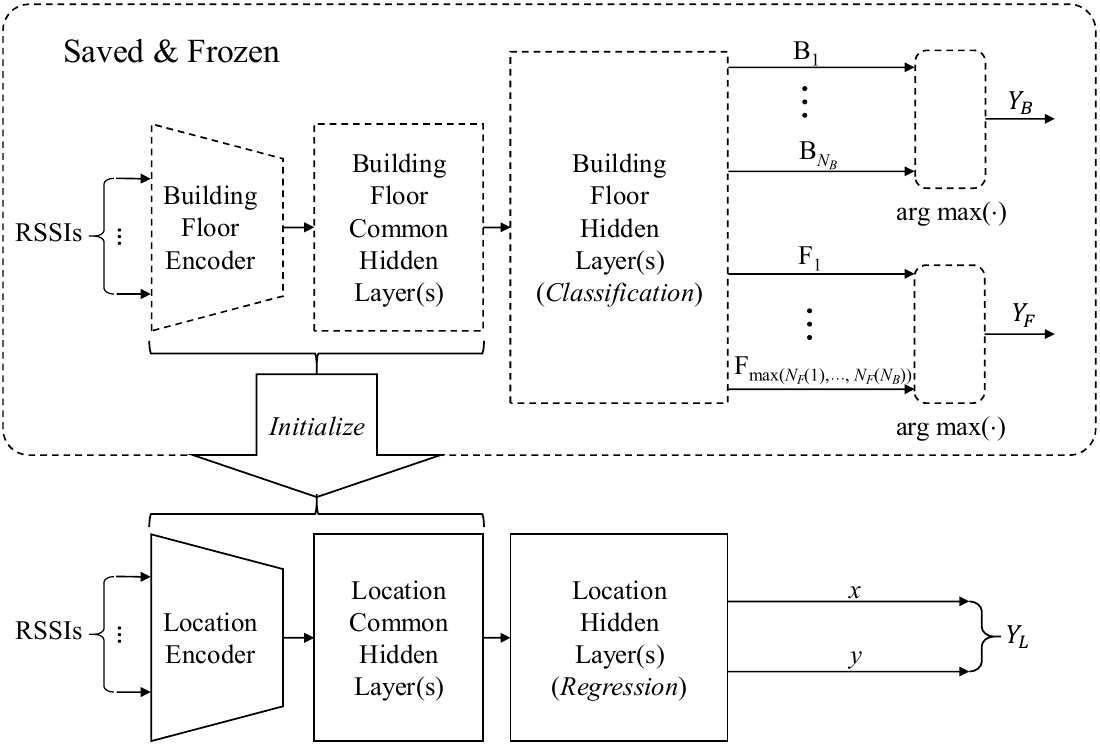}\\
    \vspace{0.3cm}%
    {\footnotesize (b) Stage 2}\\
  \end{center}
  \caption{An overview of the HST of the linked-DNN model for multi-building and
    multi-floor indoor localization.}
  \label{fig:simo_dnn_arch}
\end{figure}

Fig.~\ref{fig:simo_dnn_arch} illustrates the proposed HST of the linked-DNN
model for multi-building and multi-floor indoor localization.

Note that the hierarchical nature of the multi-building and multi-floor indoor
localization problem is fully exploited in the proposed training framework as
the DNN for the estimation of floor-level location is trained based on the
prior knowledge gained from the training of the DNN for the estimation of
building and floor while avoiding any negative impacts resulting from the
unbalanced training of the two estimation problems with the hybrid SIMO DNN in
a conventional way.

\begin{table}[!htb]
  \centering%
  \caption{Blocks of the linked-CNNLoc model shown in
    Fig.~\ref{fig:cnn_archs}~(b) for the HST framework.}
  \label{tab:exp}
  \begin{tabular}{clc}
    \toprule
    Stage                           & \multicolumn{1}{c}{Block}          & Symbol  \\
    \midrule
    \multirow{2}*{\textbf{Stage 1}} & Building Encoder                   & $E_{B}$ \\
                                    & Building Classifier                & $B$     \\
    \cmidrule(lr){2-3}
    \multirow{3}*{\textbf{Stage 2}} & Floor Encoder                      & $E_{F}$ \\
                                    & Floor 1D Convolutional Layer(s)    & $C_{F}$ \\
                                    & Floor Hidden Layer(s)              & $H_{F}$ \\
    \cmidrule(lr){2-3}
    \multirow{3}*{\textbf{Stage 3}} & Location Encoder                   & $E_{L}$ \\
                                    & Location 1D Convolutional Layer(s) & $C_{L}$ \\
                                    & Location Hidden Layer(s)           & $H_{L}$ \\
    \bottomrule
  \end{tabular}
\end{table}

Again, applying Algorithm~\ref{alg:sh} to the linked-CNNLoc model shown in
Fig.~\ref{fig:cnn_archs}~(b), based on the blocks described in
Table~\ref{tab:exp}, we can train them under the proposed HST framework as
follows:

\noindent%
\textbf{Stage 1}:
\begin{itemize}[noitemsep,topsep=0pt]
  \item Input domain: $\mathcal{X}_{1}$ is the domain of scaled RSSI values, i.e.,
        \begin{equation*}
          \begin{aligned}
            \mathcal{X}_{1} = & \big\{\left(\text{RSSI}_{1},\ldots,\text{RSSI}_{N}\right)\,\big|  \\
                              & \quad \text{RSSI}_{i}\in\mathbb{R}\text{ for }i=1,\ldots,N\big\},
          \end{aligned}
        \end{equation*}
        where $N$ is the number of APs in the building complex.
  \item Output domain: $\mathcal{Y}_{1}$ is the domain of concatenated one-hot-encoded
        categorical variables for building IDs, i.e.,
        \begin{equation*}
          \begin{aligned}
             & \mathcal{Y}_{1}=\big\{\left(b_{1},\ldots,b_{N_{B}},f_{1},\ldots,f_{N_{F}}\right)\,\big|       \\
             & \quad b_{i}\in\{0,1\}\text{ for }i=1,\ldots,N_{B}\text{ and }\sum_{i=1}^{N_{B}}b_{i}=1\big\}.
          \end{aligned}
        \end{equation*}
  \item Training dataset:
        \[
          \mathcal{T}_{1}=\left\{\left(\bm{x}_{i},\bm{y}_{i}\right)\,\middle|\,i=1,\ldots,n_{s}\right\}\subseteq\mathcal{X}_{1}\times\mathcal{Y}_{1}.
        \]
  \item Parameter set:
        \[
          \mathcal{W}_{1}=\left\{{W}_{E_{B},1},{W}_{B}\right\}.
        \]
  \item Initialization:
        \begin{align*}
           & {W}_{E_{B},1}^{init} := \text{pre-trained} \\
           & {W}_{B,1}^{init} := \text{random}.
        \end{align*}
  \item Learning algorithm:
        \[
          \mathcal{W}_{1}=\mathcal{L}\left(\mathcal{T}_{1},\mathcal{W}^{init}_{1}\right).
        \]
\end{itemize}

\noindent%
\textbf{Stage 2}:
\begin{itemize}[noitemsep,topsep=0pt]
  \item Input domain: $\mathcal{X}_{2}=\mathcal{X}_{1}$.
  \item Output domain: $\mathcal{Y}_{2}$ is the domain of floor IDs, i.e.,
        \begin{equation*}
          \begin{aligned}
             & \mathcal{Y}_{2}=\big\{\left(f_{1},\ldots,f_{N_{F}}\right)\,\big|                              \\
             & \quad f_{j}\in\{0,1\}\text{ for }j=1,\ldots,N_{F}\text{ and }\sum_{j=1}^{N_{F}}f_{j}=1\big\}.
          \end{aligned}
        \end{equation*}
  \item Training dataset:
        \[
          \mathcal{T}_{2} = \left\{\left(\bm{x}_{i},\bm{y}_{i}\right)\,\middle|\,i=1,\ldots,n_{s}\right\}\subseteq\mathcal{X}_{2}\times\mathcal{Y}_{2}.
        \]
  \item Parameter set:
        \[
          \mathcal{W}_{2}=\left\{{W}_{E_{F},2}, {W}_{C_{F},2}\,{W}_{H_{F},2}\right\}.
        \]
  \item Initialization:
        \begin{align*}
           & {W}_{E_{F},2}^{init} := {W}_{E_{B},1}, \\
           & {W}_{C_{F},2}^{init} := \text{random}, \\
           & {W}_{H_{F},2}^{init} := \text{random}.
        \end{align*}
  \item Learning algorithm:
        \[
          \mathcal{W}_{2}=\mathcal{L}\left(\mathcal{T}_{2},\mathcal{W}^{init}_{2}\right).
        \]
\end{itemize}

\noindent%
\textbf{Stage 3}:
\begin{itemize}[noitemsep,topsep=0pt]
  \item Input domain: $\mathcal{X}_{3}=\mathcal{X}_{1}$.
  \item Output domain: $\mathcal{Y}_{3}$ is the domain of scaled floor-level 2D
        coordinates, i.e.,
        \begin{equation*}
          \mathcal{Y}_{3}=\left\{(x,y)\,\middle|\, x,y \in\mathbb{R}\right\}.
        \end{equation*}
  \item Training dataset:
        \[
          \mathcal{T}_{3}=\left\{\left(\bm{x}_{i},\bm{y}_{i}\right)\,\middle|\,i=1,\ldots,n_{s}\right\}\subseteq\mathcal{X}_{3}\times\mathcal{Y}_{3}.
        \]
  \item Parameter set:
        \[
          \mathcal{W}_{3}=\left\{{W}_{E_{L},3},{W}_{C_{L},3}\,{W}_{H_{L},3}\right\}.
        \]
  \item Initialization:
        \begin{align*}
           & {W}_{E_{L},3}^{init} := {W}_{E_{F},2}, \\
           & {W}_{C_{L},3}^{init} := {W}_{C_{F},2}, \\
           & {W}_{H_{L},3}^{init} := \text{random}.
        \end{align*}
  \item Learning algorithm:
        \[
          \mathcal{W}_{3}=\mathcal{L}\left(\mathcal{T}_{3},\mathcal{W}^{init}_{3}\right).
        \]
\end{itemize}

\begin{figure}[!htb]
  \begin{center}
    \includegraphics[width=.9\linewidth]{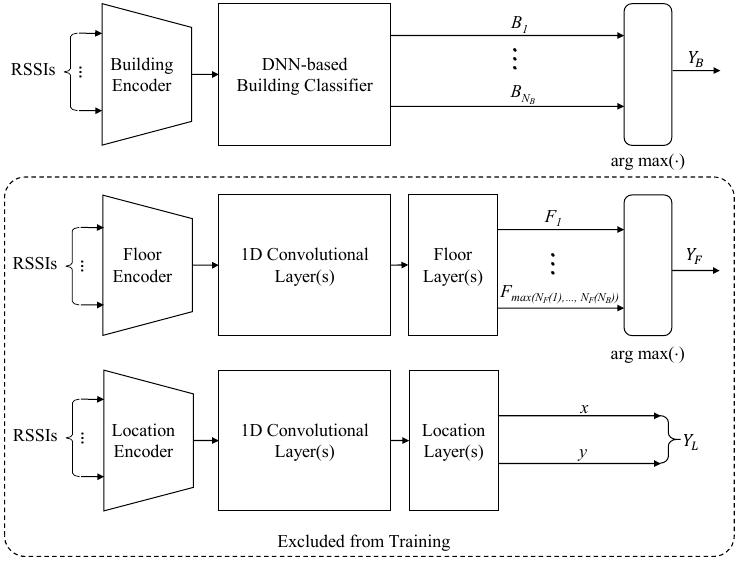}\\
    {\footnotesize (a) Stage 1}\\
    \vspace{0.2cm}
    \includegraphics[width=.9\linewidth]{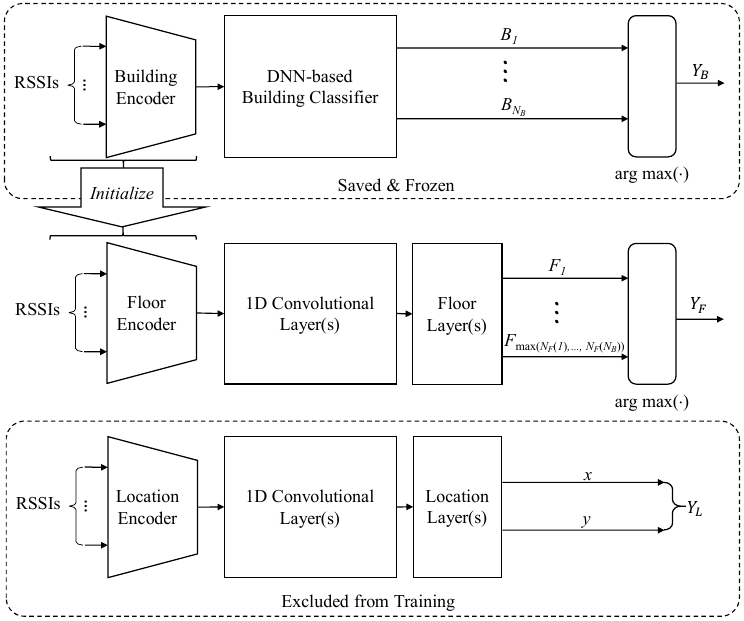}\\
    {\footnotesize (b) Stage 2}\\
    \vspace{0.2cm}
    \includegraphics[width=.9\linewidth]{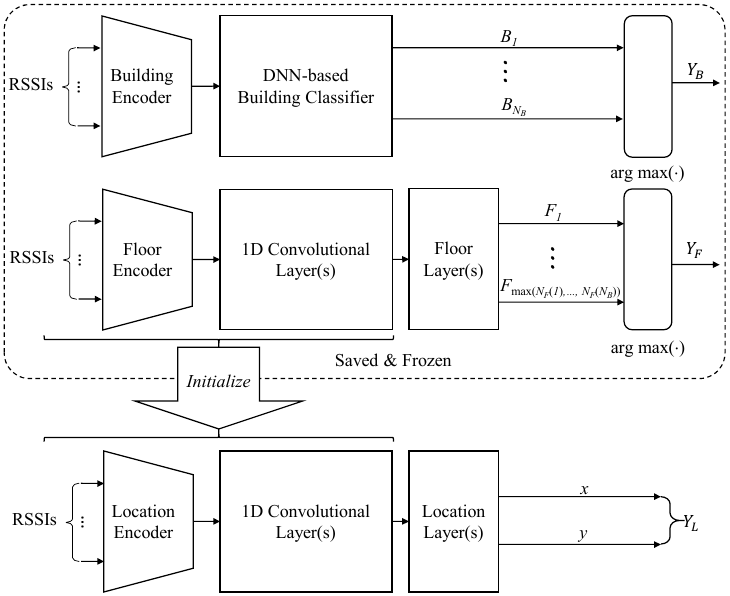}\\
    {\footnotesize (c) Stage 3}
  \end{center}
  \caption{An overview of the HST of the linked-CNNLoc model for multi-building
    and multi-floor indoor localization.}
  \label{fig:cnn_swt_arch}
\end{figure}

Fig.~\ref{fig:cnn_swt_arch} also illustrates the proposed HST of the
linked-CNNLoc model for multi-building and multi-floor indoor localization.

\section{Experimental Results}
\label{sec:exp}
To evaluate the localization performance of the linked NNs trained under the
proposed HST framework, experiments were carried out using the UJIIndoorLoc
database~\cite{rela_uji_01}. All the experiments on the NN models with the
conventional and the proposed HST frameworks were implemented using PyTorch~2.0
and Python~3.8.15 on a workstation with an Intel Core i9-9900X CPU, 128~GB RAM,
and two Nvidia GeForce RTX 2080Ti GPUs running Ubuntu 20.04 server.

The hyperparameter values for the linked-DNN model are summarized in
Table~\ref{tab:dnn}. The Adam optimizer is used with a learning rate of 0.0001
for pre-training the SAE and training the building and floor classifier and
0.001 for training the location regressor, and the batch size is set to 24.
\begin{table}[htb]
  \centering%
  \caption{Hyperparameter values for the linked-DNN model.}
  \label{tab:dnn}
  \begin{threeparttable}
    \begin{tabular}{lc}
      \toprule
      \multicolumn{1}{c}{Hyperparameter}       & Value         \\
      \midrule
      Encoder Hidden Layers                    & 520, 260, 130 \\
      Encoder Activation                       & ELU\tnote{a}  \\
      Encoder Loss Function                    & MSE           \\
      \midrule
      Common Hidden Layer                      & 520, 520      \\
      Common Hidden Layer Activation           & ELU           \\
      \midrule Building and Floor Hidden Layer & 8             \\
      Building and Floor Output Layer
      Activation                               & Sigmoid       \\
      Building and Floor Loss Function         & BCE\tnote{b}  \\
      \midrule Location Hidden Layers          & 520, 2        \\
      Location Hidden Layers Activation        & Tanh          \\
      Location Output Layers Activation        & Linear        \\
      Location Loss Function                   & MSE           \\
      \bottomrule
    \end{tabular}
    \begin{tablenotes}
      \item[a] Exponential linear unit.
      \item[b] Binary cross entropy.
    \end{tablenotes}
  \end{threeparttable}
\end{table}

As for the linked-CNNLoc model, we apply the Adam optimizer with a fixed
learning rate of 0.0001 across all sub-models, including SAE. Note that during
the classifier and estimator training, learning rate schedulers are used,
decreasing the learning rate by a factor of 0.1 with patience of 5 for the
floor and a factor of 0.5 with the same patience for the location and the batch
size of 26. The detailed hyperparameter settings are summarized in
Table~\ref{tab:cnn}.
\begin{table}[!htb]
  \centering%
  \caption{Hyperparameter values for the linked-CNNLoc model.}
  \label{tab:cnn}
  \begin{threeparttable}
    \begin{tabular}{lc}
      \toprule
      \multicolumn{1}{c}{Hyperparameter} & Value               \\
      \midrule
      Encoder Hidden Layers              & 520, 260, 130       \\
      Encoder Activation                 & ELU                 \\
      Encoder Loss Function              & MSE                 \\
      \midrule
      Building Hidden Layers             & 130, 130, 3         \\
      Building Hidden Layers Activation  & ELU                 \\
      Building Output Layers Activation  & softmax             \\
      Building Loss Function             & CE\tnote{a}         \\
      \midrule
      Convolutional Layers               & 99-22, 66-22, 33-22 \\
      Convolutional Layers Activation    & ELU                 \\
      \midrule
      Floor Output Layer                 & 5                   \\
      Floor Output Layers Activation     & softmax             \\
      Floor Loss Function                & CE                  \\
      \midrule
      Location Output Layer              & 2                   \\
      Location Output Layers Activation  & Linear              \\
      Location Loss Function             & MSE                 \\
      \bottomrule
    \end{tabular}
    \begin{tablenotes}
      \item[a] Cross entropy.
    \end{tablenotes}
  \end{threeparttable}
\end{table}

Table~\ref{tab:reference-vs-proposed} summarizes the multi-building and
multi-floor indoor localization performance and 3D localization error, which
are defined in~\cite{intro_art_05}, of the reference models trained in a
conventional way and the proposed models based on linked NNs trained under the
HST framework.
\begin{table*}[!htb]
  \centering%
  \caption{The multi-building and multi-floor indoor localization performance of the
    reference and proposed models.}
  \label{tab:reference-vs-proposed}
  \begin{tabular}{cccccccc}
    \toprule
    \multirow{2}{*}{Model} & \multirow{2}{*}{Building hit rate} & \multirow{2}{*}{Floor hit rate} & \multicolumn{5}{c}{3D error}                                                                   \\ \cmidrule{4-8}
                           &                                    &                                 & Average                      & Std.          & Min.          & Median        & Max.            \\ \midrule
    Reference DNN          & 100\%                              & 93.34\%                         & \SI{8.45}{\m}                & \SI{8.61}{\m} & \SI{0.31}{\m} & \SI{6.14}{\m} & \SI{130.55}{\m} \\
    Reference CNN          & 100\%                              & 92.89\%                         & \SI{9.10}{\m}                & \SI{8.06}{\m} & \SI{0.36}{\m} & \SI{6.74}{\m} & \SI{65.18}{\m}  \\
    \textbf{Proposed DNN}  & 100\%                              & 93.34\%                         & \SI{8.19}{\m}                & \SI{7.55}{\m} & \SI{0.18}{\m} & \SI{5.94}{\m} & \SI{79.32}{\m}  \\
    \textbf{Proposed CNN}  & 100\%                              & 92.80\%                         & \SI{8.71}{\m}                & \SI{7.74}{\m} & \SI{0.19}{\m} & \SI{6.45}{\m} & \SI{60.69}{\m}  \\
    \bottomrule
  \end{tabular}
\end{table*}

From the table, we can observe that the proposed models based on linked NNs
trained under the HST framework outperform the reference models trained under
the conventional training method in multi-building and multi-floor indoor
localization. Specifically, the average 3D errors of the proposed models are
significantly lower than those of the reference models, i.e., \SI{8.19}{\meter}
vs. \SI{8.45}{\meter} for the DNN models and \SI{8.71}{\meter} vs.
\SI{9.10}{\meter} for the CNN models. The overall statistics of 3D error of the
proposed models are also improved compared to the reference models: For
example, the maximum 3D localization errors of the proposed models are much
smaller than those of the reference models, i.e., \SI{79.32}{\meter} vs.
\SI{130.55}{\meter} for the DNN models and \SI{60.69}{\meter} vs.
\SI{65.18}{\meter} for the CNN models. Compared to 3D errors, the building hit
rates are 100\% for all the models, and there is a slight decrease in the floor
hit rate of the proposed CNN model (i.e., from 92.89\% to 92.80\%).

We also compare the multi-building and multi-floor indoor localization
performance of the proposed models with that of the state-of-the-art DNN-based
models in Table~\ref{tab:comp-sota}, which shows that the proposed DNN model
provides the smallest 3D error; to the best of the authors' knowledge, this is
the most accurate result obtained for the whole of the UJIIndoorLoc database
based on DNN-based models.
\begin{table}[!htb]
  \centering%
  \caption{The multi-building and multi-floor indoor localization performance
    of the state-of-the-art models.}
  \label{tab:comp-sota}
  \begin{threeparttable}
    \begin{tabular}{lccc}
      \toprule
      \multicolumn{1}{c}{Model}            & Building hit rate & Floor hit rate   & 3D error           \\
      \midrule
      Simple DNN~\cite{intro_art_02}       & N/A               & 92.00\%\tnote{a} & N/A                \\
      Scalable DNN~\cite{intro_art_04}     & 99.82\%           & 91.27\%          & \SI{9.29}{\meter}  \\
      Fusion~\cite{rela_stc_02}            & N/A               & 95.13\%\tnote{a} & N/A                \\
      Hierarchical RNN~\cite{rela_rnn_01}  & 100\%             & 95.24\%          & \SI{8.62}{\meter}  \\
      CNNLoc~\cite{rela_cnn_01}\tnote{b}   & 100\%             & 96.03\%          & \SI{11.78}{\meter} \\
      MOGP RNN~\cite{rela_rnn_02}\tnote{b} & 100\%             & 94.20\%          & \SI{8.42}{\meter}  \\
      \textbf{Proposed DNN}                & 100\%             & 93.34\%          & \SI{8.19}{\meter}  \\
      \textbf{Proposed CNN}                & 100\%             & 92.80\%          & \SI{8.71}{\meter}  \\
      \bottomrule
    \end{tabular}
    \begin{tablenotes}
      \item[a] Joint building/floor hit rate.
      \item[b] The algorithms involve data augmentation.
    \end{tablenotes}
  \end{threeparttable}
\end{table}

Note that both proposed models have lower floor hit rates than the hierarchical
RNN and CNNLoc models. The CNNLoc model, however, relies on complex augmented
data, while the hierarchical RNN is based on the RNN architecture that is
computationally slower than other NN architectures due to its recursive nature.

The computational advantage of the proposed models over the hierarchical RNN is
clearly shown in Table~\ref{tab:time}, which compares the training times of the
models on the same workstation.
\begin{table}[!htb]
  \centering%
  \caption{Model training times over five independent runs.}
  \label{tab:time}
  \begin{tabular}{lc}
    \toprule
    \multicolumn{1}{c}{Model}           & Average time [\SI{}{\minute}] \\
    \midrule
    Hierarchical RNN~\cite{rela_rnn_01} & 7.551                         \\
    Reference DNN                       & 2.176                         \\
    Reference CNN                       & 3.199                         \\
    \textbf{Proposed DNN}               & 2.598                         \\
    \textbf{Proposed CNN}               & 3.381                         \\
    \bottomrule
  \end{tabular}
\end{table}

According to Table~\ref{tab:time}, the training times of the proposed models
under the HST framework are slightly higher than those of the reference models
but much lower than that of the hierarchical RNN~\cite{rela_rnn_01} under a
conventional training framework, which is quite remarkable considering that
multiple linked NNs of the proposed models are trained through two or three
stages under the HST framework.

\section{Conclusions}
\label{sec:conc}
A new solution to the problem of large-scale multi-building and multi-floor
indoor localization based on linked NNs trained under the HST framework has
been proposed in this paper, which exploits the hierarchical nature of the
location information of Wi-Fi fingerprints. The HST framework further extends
the stage-wise training framework~\cite{intro_art_06} to the case of multiple
linked NNs to efficiently handle the problem of large-scale multi-building and
multi-floor indoor localization by training the model for the estimation of
floor-level location (i.e., a lower-level sub-problem at a later stage) based
on the prior knowledge gained from the training of the model(s) for the
estimation of building and floor IDs (i.e., a higher-level sub-problem at an
earlier stage). The linked-DNN and linked-CNNLoc models are also derived for
the HST framework based on the modified versions of hybrid SIMO
DNN~\cite{rela_dnn_01} and CNNLoc~\cite{rela_cnn_01}, the latter of which are
taken as reference models for the conventional training framework.

The experimental results with the UJIIndoorLoc multi-building and multi-floor
fingerprint database demonstrate that the proposed models based on linked NNs
trained under the HST framework outperform the reference models trained in a
conventional way in multi-building and multi-floor indoor localization.
Specifically, the linked-DNN model results in a 3D localization error of
\SI{8.19}{\meter}, which, to the best of the authors' knowledge, is the best
result obtained for the whole of the UJIIndoorLoc database based on any DNN
(including CNN and RNN) models. The linked-CNNLoc model also outperforms the
original CNNLoc model (i.e., \SI{8.71}{\meter} vs. \SI{11.78}{\meter}) with the
same 1D-CNN parameter settings and without any additional data augmentation.

Note that the results with both SIMO DNN and CNNLoc models presented in this
paper demonstrate that the proposed HST framework can be applied to different
NN architectures and extended to other sources of sensed data with hierarchical
structures, such as fingerprints from other wireless signal sensors (including
but not limiting to BLE, Zigbee, and LoRaWAN), to further improve the
localization performance in an indoor environment.

\balance 


\end{document}